\documentclass{article}

\usepackage{arxiv}

\usepackage[utf8]{inputenc} 
\usepackage[T1]{fontenc}    
\usepackage{hyperref}       
\usepackage{url}            
\usepackage{booktabs}       
\usepackage{amsfonts}       
\usepackage{nicefrac}       
\usepackage{microtype}      
\usepackage{lipsum}		
\usepackage{graphicx}
\usepackage{natbib}
\usepackage{doi}

\usepackage{amsmath}
\usepackage{multirow}
\usepackage{subfig}

\title{Automatic Language Identification for Celtic Texts}

\date{} 					

\author{ \href{https://orcid.org/0000-0002-3122-2728}{\includegraphics[scale=0.06]{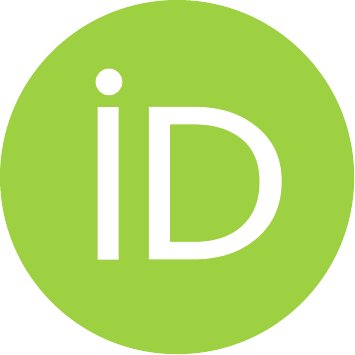}\hspace{1mm}Olha Dovbnia}\\
	Warsaw University of Technology\\
	Warsaw, Poland \\
	\texttt{olgadovbnia97@gmail.com} \\
	\And
	\href{https://orcid.org/0000-0002-3407-7570}{\includegraphics[scale=0.06]{orcid.pdf}\hspace{1mm}Anna Wróblewska} 
	\\
	Warsaw University of Technology\\
	Warsaw, Poland \\
	\texttt{anna.wroblewska1@pw.edu.pl} \\
}

\renewcommand{\shorttitle}{Automatic Language Identification for Celtic Texts}

\hypersetup{
pdftitle={Automatic Language Identification for Celtic Texts},
pdfsubject={CS.CL, cs.LG},
pdfauthor={Olha Dovbnia, Anna Wróblewska},
pdfkeywords={Natural language identification, Natural language processing, Low-resource languages, Machine learning, Classification},
}

\begin{document}
\maketitle

\begin{abstract}
Language identification is an important Natural Language Processing task. It has been thoroughly researched in the literature. However, some issues are still open. 
This work addresses the identification of the related low-resource languages on the example of the Celtic language family.

This work's main goals were: (1) to collect the dataset of three Celtic languages; (2) to prepare a method to identify the languages from the Celtic family, i.e. to train a successful classification model; (3) to evaluate the influence of different feature extraction methods, and explore the applicability of the unsupervised models as a feature extraction technique; (4) to experiment with the unsupervised feature extraction on a reduced annotated set. 

The Celtic language group is not well researched, so, in this work, we collected a new dataset as no dataset for the combination of the selected languages was available. The dataset includes Irish, Scottish, Welsh and English records.

Language identification is a classification task, so supervised models such as SVM and neural networks were applied. Traditional statistical features were tested alongside the output of clustering, autoencoder, and topic modelling methods.
The analysis showed that the unsupervised features could serve as a valuable extension to the n-gram feature vectors. It led to an improvement in performance for more entangled classes. The best model achieved 98\% F1 score and 97\% MCC. The dense neural network consistently outperformed the SVM model.

The low-resource languages are also challenging due to the scarcity of the available annotated training data. This work evaluated the performance of the classifiers using the unsupervised feature extraction on the reduced labelled dataset to handle this issue. The results uncovered that the unsupervised feature vectors are more robust to the labelled set reduction. Therefore, they proved to help to achieve comparable classification performance with much less labelled data.
\end{abstract}


\section{Introduction}

Language Identification (LI) approaches the problem of automatic recognition of specific natural languages~\citep{survey}.
LI has multiple applications in the modern world. The original use case is routing source documents to language-appropriate Natural Language Processing (NLP) components, such as machine translation and dialogue systems~\citep{zhang-etal-2018-fast-compact}. Most of the other NLP methods assume input text is monolingual; in that case, routing of the input documents is crucial. Moreover, LI is a valuable component of the corpus creation pipelines, especially for the low-resource languages. Text crawled from the web resources usually needs to be separated by language to create a meaningful corpus.

LI can be a relatively easy task for the most popular languages with the abundance of available resources. Nevertheless, there are languages with fewer native speakers and less research focus; such languages are considered low-resource. Low-resource languages and related languages continuously pose a problem for language identification~\citep{he-etal-2018-discriminating, survey}. This paper focuses on applying language identification methods to a family of low-resource languages on the example of the Celtic language group combining the two issues.

The main problem with the low-resource languages is the unavailability of high-quality corpora, which is costly to prepare. Language Identification faces this problem same as other NLP tasks. Thus, in our research, we created a corpus of three Celtic languages. It contains Irish, Scottish, and Welsh texts. Moreover, the corpus is extended with a small proportion of English samples because these languages are usually mixed in the common usage. 

This work also focuses on the preparation of a language identification model alongside the evaluation and analysis of the different feature extraction methods. In particular, we explored the possibility of the application of the output of the unsupervised learning models as a feature representation for the classification task. The unsupervised approach was promising in the case of scarce annotated data as here. The unsupervised models operate on the unlabelled data and so enhance the classification model's performance on a limited labelled dataset. To solve the lack of large corpora for low-resource languages, we proved that the features extracted from unsupervised methods ensure high performance on the reduced labelled set size. Consequently, this study validates that it is possible to save resources on data annotation, determine the language class of low-resource languages more efficiently, and ensure that the data for further processing is of high quality.

In this study, our main contributions are: 
\begin{enumerate}
\item collecting the dataset of three Celtic languages (Irish, Scottish, Welsh) and English (Section~\ref{sec:dataset}),
\item preparing a method to identify these languages from the Celtic family and differentiate them from English (Section~\ref{sec:approach}),
\item evaluating the influence of different feature extraction methods and exploring the applicability of unsupervised models as a feature extraction technique (Section~\ref{sec:results}),
\item experimenting with unsupervised feature extraction on a reduced annotated set (Section~\ref{sec:reduced}).
\end{enumerate}

In the following sections, we discussed the related research (Section~\ref{sec:related-works}), the dataset creation process (Section~\ref{sec:dataset}), the proposed approach (Section~\ref{sec:approach}), the results of the experiments (Section~\ref{sec:results}), the list of conclusions (Section~\ref{sec:conclusion}) and directions for future research (Section~\ref{sec:future-work}).

\section{Related Work}
\label{sec:related-works}

Language Identification faces unique challenges compared to other NLP tasks. Most NLP problems assume that the language of the source data is known. Therefore it is possible to apply language-specific approaches to tokenisation and grammar representation. Moreover, the text belonging to the same language can be written with varying orthography and encodings. It leads to a broader language class definition for LI, meaning LI must classify samples to the same language regardless of the orthography and encoding differences.

The most popular features in LI literature are n-grams both on character and word level \citep{hanani-etal-2017-identifying, vatanen-etal-2010-language}. While most of the articles address them in some way, more unusual features appear as well. \citet{van-der-lee-van-den-bosch-2017-exploring} enriched n-gram classification with text statistics and POS-tagging to differentiate varieties of the Dutch language. 

In the existing literature, we found no research on the identification of the Celtic languages with the assumption that samples are monolingual. 
Table~\ref{similar-results} presents a summary of articles working on related low-resource languages.

\begin{table}[!ht]
\centering
\caption{Results from research papers on similar tasks -- related low-resource languages identification. Note: char LM -- character language model}
\begin{tabular}{ p{3.2cm}p{2.5cm}lp{1.2cm}cc } 
 \hline
 \textbf{Article} & \textbf{Languages} & \textbf{\# of Classes} & \textbf{Classifier} & \textbf{Accuracy} & \textbf{F1} \\ 
 \hline
 \citet{minocha-tyers-2014-subsegmental} & Irish, Welsh, Breton & 3 pairs & Char LM & 74-89\% & 32-63\% \\
 \citet{philippines} & Philippines local languages, English & 8 & SVM & 92\% & 97\% \\
 \citet{he-etal-2018-discriminating} & Uyghur, Kazakh & 2 & MaxEnt & 96\% & 95\% \\
 \citet{palakodety-khudabukhsh-2020-annotation} & Indian local languages & 9 & Weak Labels & 98\% & 98\% \\
 \hline
\end{tabular}
\label{similar-results}
\end{table}

\citet{minocha-tyers-2014-subsegmental} dealt with Irish and Welsh languages. However, they had a bit different task compared to us resulting in the different dataset and the application of other models. They attempted to perform language identification on code-switching samples, i.e., to identify segments of different languages in one sample. Their classifiers performed the binary classification on language pairs of a Celtic language and English. The dataset collected and used in their work was scarce (40 to 50 samples per language pair), so authors had to use features based on word lists.

\citet{philippines} performed classification of the Philippines languages without code-switching. Their dataset had similar class sizes to the ones chosen in our work, but the number of classes was higher. Additionally, the level of closeness between the classified languages is not known and several language groups can be included.

Another paper addresses the separation of the Uyghur and Kazakh short texts \citep{he-etal-2018-discriminating}. These languages are also low-resource and closely-connected. The similarities between these languages at the word level are about 90\%. This is a much higher similarity compared to the Celtic languages. A large corpus was created and professionally annotated in this work ensuring high quality of the data.

Our setting was different from these articles due to either the dataset size and quality, languages' closeness or the task's goal.

Recent LI surveys~\citep{hughes-etal-2006-reconsidering} and~\citep{survey} outlined research areas yet to be investigated including multilingual text~\citep{al-badrashiny-diab-2016-lili},~\citep{mandal-singh-2018-language}; related languages \citep{similar-lang}, \citep{acs-etal-2015-two}; low-resource languages \citep{south-african2}, \citep{minocha-tyers-2014-subsegmental}, \citep{philippines}; unsupervised LI \citep{lin-etal-2014-cmu}, \citep{twitter-short}. 

In our research, the task of LI was performed on languages from the Celtic family. The whole language family has a deficient number of native speakers making this language group low-resource. Additionally, we applied unsupervised models as feature extraction methods to explore the possibility of reducing the need in an extensive labelled dataset (using a large amount of unlabelled data with a smaller proportion of the labelled entries). Therefore, we addressed the issues of the related low-resource languages and unsupervised LI.

\section{Collected Dataset}
\label{sec:dataset}

\begin{figure}[!ht]
\centering
\includegraphics[scale=0.9]{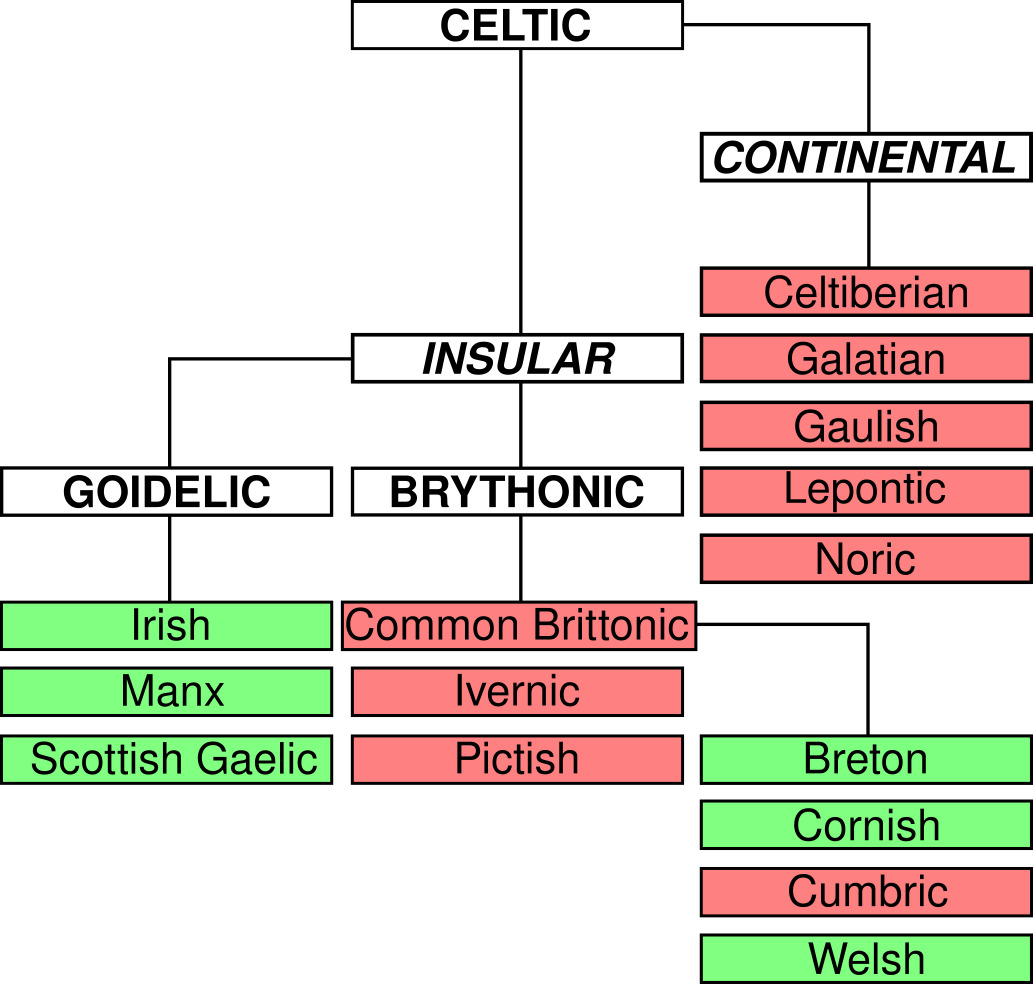}
\caption{Celtic languages family tree (dark grey -- extinct; light grey -- alive). Source:~\cite{celtic-tree}}
\label{celtictree}
\end{figure}

We considered the most popular Celtic languages: Irish, Scottish Gaelic, and Welsh. A tree in Figure~\ref{celtictree} summarises the relationship between them. Irish and Scottish belong to the same language subgroup. Consequently, they are likely to be more challenging to distinguish. English, on the other hand, comes from a completely different language family.

The corpus for a combination of these Celtic languages had not existed. In this work, it was aggregated from resources available online for each of the languages. The process of the collection and preparation of the language labels for each language is described below.

\subsection{Irish}
    
Irish data comes from the collection of historical texts (up to 1926)~\citep{irish-hist} which was the only open corpus of Irish that we had found. The data was not prepared for immediate processing and required scraping from the web page. 
    
We divided the text into sentences and filtered by language using a FastText model~\citep{joulin2016fasttext}, \citep{joulin2016bag}, \citep{fasttext-model} as some English sentences appeared in the data. 
English sentences were filtered out according to a threshold set experimentally to 0.5. 
    
\subsection{Scottish Gaelic}
    
The source corpus for Scottish Gaelic~\citep{scottish-corpus} was annotated at code-switching points of Scottish and English. It required scraping and sentencing. During sentencing, an additional category of code-mixed text emerged and was filtered out. 
    
\subsection{Welsh}
    
Publicly available language resources for Welsh were the most scarce. Three sources were combined: raw data from research by~\citet{welsh-corpus}; dataset of Welsh phrases~\citep{welsh-phrases}; historical texts \citep{welsh-hist}. After sentencing, the same FastText annotation as for Irish was applied with a new experimental threshold set to 0.8. 
    
\subsection{English}
    
English represents a language class that is commonly mixed with these Celtic languages. English sentences were taken from a reduced version of British National Corpus~\citep{english-bnc}. Additional quality checks were not necessary in this case.

At last, a final dataset was compiled. 
Class sizes were chosen similar to those used by~\citet{philippines}. The authors of this article performed LI on the related low-resource languages similar to our task. So we selected such class sizes to guarantee the new dataset will be sufficient for the task.
Detailed corpus statistics are presented in the Table~\ref{dataset-table}.

\begin{table}
\centering
\caption{Final dataset statistics}
\begin{tabular}{ rccccc } 
 \hline
 & \textbf{Total} & \textbf{Irish} & \textbf{Scottish} & \textbf{Welsh} & \textbf{English} \\ 
 \hline
 Sentences & 9,969 & 2,689 & 2,582 & 3,098 & 1,600 \\ 
 Unique sentences & 9,942 & 2,688 & 2,568 & 3,086 & 1,600 \\ 
 Words & 172,655 & 46,495 & 52,745 & 48,688 & 24,727 \\
 Average words/sentence & 17.31 & 17.29 & 20.42 & 15.71 & 15.45 \\
 \hline
\end{tabular}
\label{dataset-table}
\end{table}

Tables \ref{dataset-irish}, \ref{dataset-scottish}, \ref{dataset-welsh} provide examples of records in the final dataset for each of the Celtic languages. These examples can give the general idea what kind of data we were working with.

\begin{table}[ht]
\centering
\begin{tabular}{ c } 
 \hline
 Ag Napoli bhí marcaidheacht bhreá againn suas go mainistir Shan Martino atá suidhte ar \\ bhárr cnuic mhóir a bhfuil radharc áluinn uaidh ar an gcathair na hiomlán thíos faoi. \\
 \hline
 Is iongantach liom gur chreidis an chaint sin, agus tá a fhios agamsa cad  \\ na thaobh gur cuireadh an tuairisc sin chugat. \\
 \hline
 Dá ndeintí an méid sin badh mhór an áise a dhéanfadh sé agus ní chloisfidhe \\ a leathoiread cnáimhseála. \\
 \hline
 Chun an léightheora annso síos. \\
 \hline
 Badh mhian liom, ar Draoidheantóir go gcluinfinn a cheol má tá binn. \\
 \hline
\end{tabular}
\caption{Examples of Irish dataset records}
\label{dataset-irish}
\end{table}

\begin{table}[ht]
\centering
\begin{tabular}{ c } 
 \hline
 Tha an luach air a phàigheadh, agus tha sinne ann an gabhail ri sin air ar deanamh \\ n ar saor-dhaoine de n bhaile. \\
 \hline
 Coma leibhse có dhiù, a Lachuinn, tha chùis a dol gu math mur faod mi éisdeachd \\ ri spleadhraich Sheòruis cho math ribhse. \\
 \hline
 Mata tha feum air féin-cheasnachadh ann. \\
 \hline
 Ach ciod a bhuaidh a dh fhag i air cho beag de fhocail; agus gu h-araid ciod a bhuaidh \\ a thug dhi cho beag de fhocail ghoirid an coimeas ris na bheil innte de fhocail fhada? \\
 \hline
 A rádh gur béigean dó imtheacht agus ná feicfinn airís é! \\
 \hline
\end{tabular}
\caption{Examples of Scottish dataset records}
\label{dataset-scottish}
\end{table}

\begin{table}[ht]
\centering
\begin{tabular}{ c } 
 \hline
 Fe welwch or llun fod dau soced ar gyfer trydan y garafan. \\
 \hline
 Trechwyd bwriadaur swyddogion ar 25 Mai, ond ymhen deuddydd cyrchwyd yr ardal gan \\ ugain o blismyn, a chael fod 300 o bobl yn eu disgwyl yn y Ty Nant Inn. \\
 \hline
 Arol denu sylw fel cantores addawol, fe ddechreuodd ei band cyntpan oedd hin ddeuddeg oed. \\
 \hline
 Dyna maen debyg fydd disgwyliadau Aelodaur Cynulliad hefyd. \\
 \hline
 Mae gan Gwyn Thomas gerdd drawiadol yn darlunio carcharor gwleidyddol yn \\ cadw ei bwyll trwy ymgyfeillachu a chrocrotsien yn unigrwydd ei gell. \\
 \hline
\end{tabular}
\caption{Examples of Welsh dataset records}
\label{dataset-welsh}
\end{table}

\section{Our Approach}
\label{sec:approach}

LI is a classification task, so supervised models should be applied. Nevertheless, these models need a large dataset to achieve better results. This consideration becomes an issue when working with low-resource languages as there are no large high-quality datasets available. As a labelled dataset for classification models is costly to compile, we explored the possibility to utilize the advantages of the unsupervised learning in the LI pipeline. 

Semi-supervised feature learning allowed combining the benefits of these two approaches. An unsupervised model identified the hidden patterns in the whole dataset and transformed the data into a latent representation. Later, the created structural information was employed as the features for the classification models.

We tested the following machine learning models: a support vector machine (SVM), a dense neural network (NN), and a convolutional neural network (CNN) as the classifiers operating on the features described below.

\subsection{Preprocessing}
The same preprocessing was applied in all our experiments. It involved the removal of punctuation, special symbols, digits and the application of lowercasing. During the experiments, 80\% of the dataset was taken as the training set and 20\% as the testing set. We used a random constant train-test split of our dataset in all experiments.

\subsection{Characters}
The first feature type that was tested was the characters. The intuition behind this choice was to preserve the sequential characteristics of the input data. Each character was encoded to an integer. 
For this kind of feature representation, we expected meaningful results from the deep neural models -- CNN. 

\subsection{Character N-grams}
N-grams are one of the most popular features mentioned in the LI literature \citep{hanani-etal-2017-identifying, vatanen-etal-2010-language}.
Character n-grams were chosen instead of word-level features due to the similarities between the Celtic languages and the possible word overlap.
This and all the following feature extraction methods were tested on the SVM and NN models.

\subsection{Text Statistics}
Text statistics served as an extension for the n-gram features.
We tested the impact of combining the average word length and the average number of consonants per word. 

\subsection{Clustering}
The next features utilized unsupervised models' output. 

Clustering attempts to identify natural groups within the input data.
We created k-means, Birch (Balanced Iterative Reducing and Clustering using Hierarchies), agglomerative clustering, and Gaussian Mixture models based on the features described above. 
A variety of the clustering methods was expected to provide a fine-grained combination of the identified clusters. The perspective of the different models was expected to be able to reflect the actual distribution of the language classes. When testing this feature, the language classifier takes in a vector combining the group assignments from the all four clustering approaches.

\subsection{Variational Autoencoder}
\begin{figure}
\centering
\includegraphics[scale=0.3]{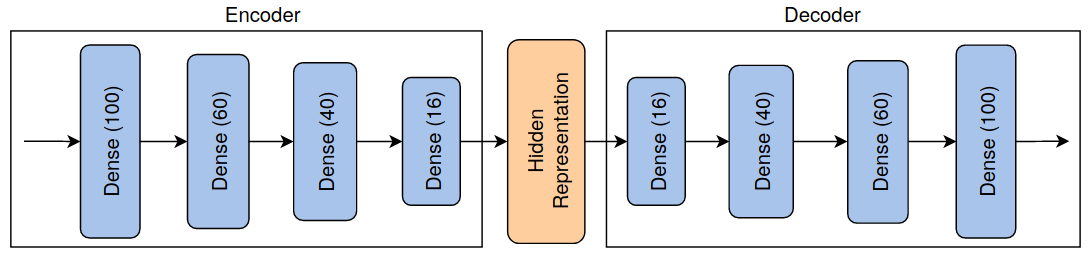}
\caption{VAE architecture used in our experiments. Numbers in parentheses stand for the number of neurons in a layer}
\label{vae-arch}
\end{figure}

Variational autoencoder (VAE) is an unsupervised generative model.
The encoder is forced to reduce the input's dimensionality while preserving the most important information for the decoder. The model produces a hidden representation suitable for application in a classification model. The output of the latent layer was used as a feature representation for language classifiers.
VAE architecture used during our experiments took in single characters from the input data to the fully-connected encoder and decoder (Figure~\ref{vae-arch}). The hidden representation was 2-dimensional. The architecture was chosen empirically.

\subsection{Topic Modeling}
Another way to utilize unsupervised models for feature learning from the text is topic modeling. This approach creates word clusters from a collection of documents representing the distinct topics. One sentence may contain several topics. This fact provides greater flexibility when working with a set of related languages by addressing the possible word overlap. The topic modelling approach is useful in finding common words for all languages and identifying the specific words for each of them. While clustering, in our use case, treats one sample as a whole and the classifier can be misled by common words within the sample. When using the topic modelling features, the classifier deals with more fine-grained word groups and potentially can perform the language identification better.

One of the most popular topic modeling methods is Latent Dirichlet Analysis (LDA). 
The number of topics was set to four -- the same as the number of languages in the corpus. 
At the prediction stage, LDA returned a vector of each topic's respective proportions. We decided to preserve this full amount of the information for the classifier.
LDA feature vector contained the probabilities of the input data belonging to each of the topics.

\section{Data Insights from the Unsupervised Models}\label{insights}
To get a better understanding of the relationships in the multilingual data, we can apply the unsupervised learning approaches. The three tested unsupervised methods are based on different features, so they are expected to reflect the problem from the different perspectives. 

\subsection{Clustering}
\begin{figure}[!ht]
\centering
\includegraphics[scale=0.32]{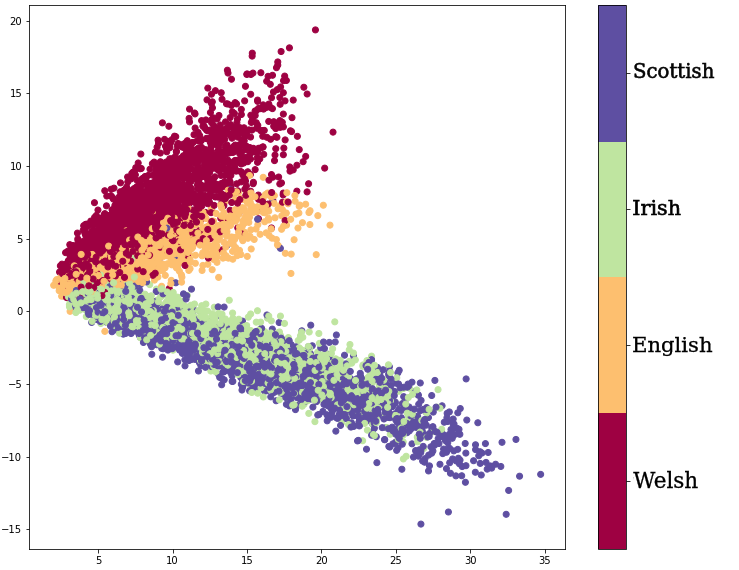}
\caption{2D statistical features representation: n-grams and text statistics. The features forms natural clusters for the four considered languages, however the distinction between Scottish and Irish is more challenging}
\label{simple-feats}
\end{figure}

Clustering models were trained based on the statistical features such as n-grams and text statistics (average word length, average number of consonants per word). 
To get an idea how the statistical features are represent the natural-language classes, the statistical feature space was transformed to a 2-dimensional representation using the Principal Component Analysis (PCA) (see Figure~\ref{simple-feats}). 
With this test we discovered the Welsh has the closest connection to English, but these languages remain mostly separable from each other. Moreover, they are independent from the Irish-Scottish pair. 
Distinguishing between Irish and Scottish appears to be a more demanding problem as they are not linearly separable in this space; however, they are distinct from Welsh and English.

\begin{figure}
 \centering
 \subfloat[Birch]{\includegraphics[scale=0.25]{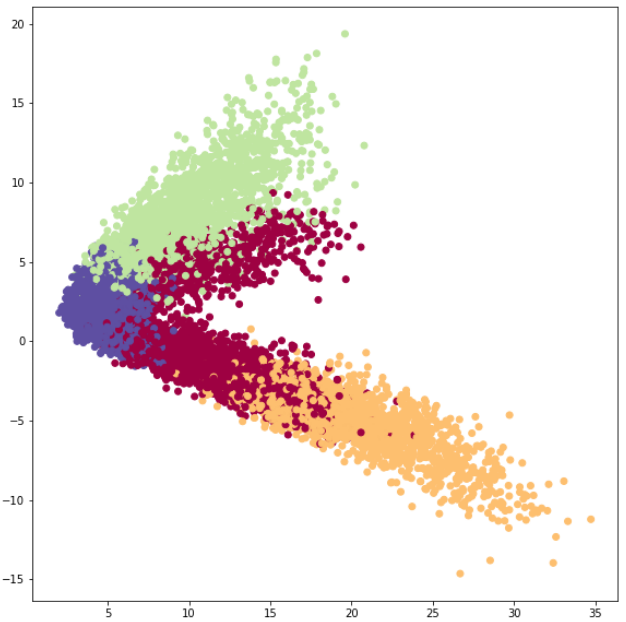}\label{birch}}%
 \hfil
 \subfloat[Agglomerative]{\includegraphics[scale=0.25]{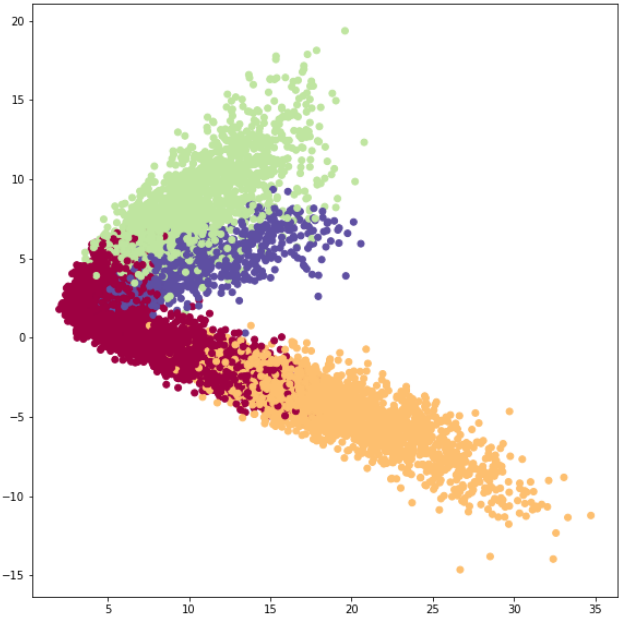}\label{aggl}}%
 \hfil
 \subfloat[K-means]{\includegraphics[scale=0.25]{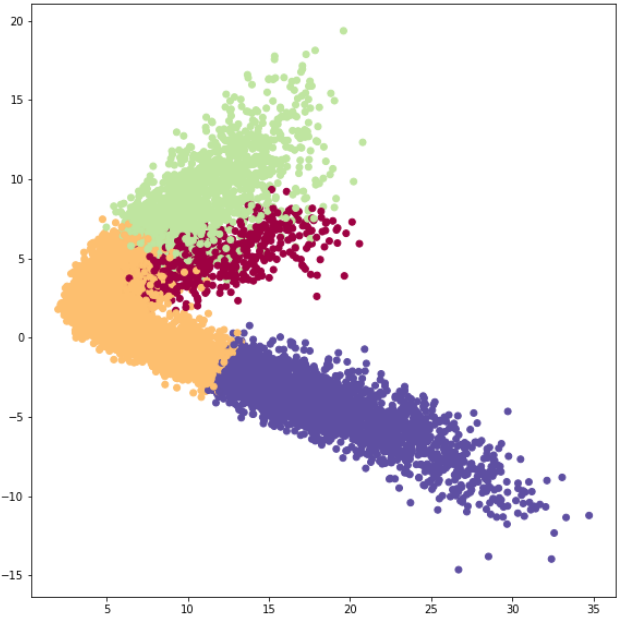}\label{kmeans}}%
 \hfil
 \subfloat[Gaussian Mixture]{\includegraphics[scale=0.25]{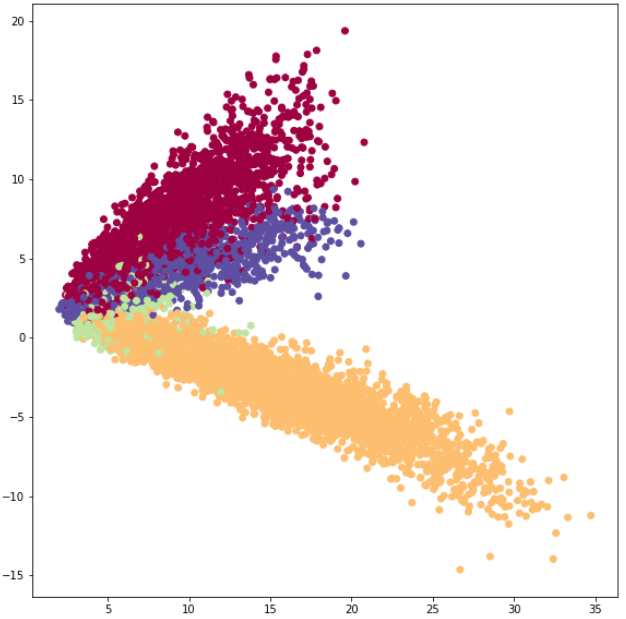}\label{gm}}%
 \caption{Clusters found by the clustering algorithms on the same feature space and PCA transformation of our dataset as in the Figure~\ref{simple-feats}. The 4 shades of grey represent the identified groups without any language mapping}
 \label{clusters-graph}
\end{figure}

Plots~\ref{birch} and~\ref{aggl} show the clusters found by the hierarchical clustering models. Both methods identified the clusters vaguely corresponding to Welsh and English. However, the two other languages proved to be challenging. 

The Gaussian mixture provides the best-fitted representation (Figure~\ref{gm}) among all clustering methods. This model makes errors only with a cluster mapping to the Irish language wrongly classifying most of its samples to Scottish. K-means algorithm (Figure~\ref{kmeans}) has identified clusters with similar distribution as the agglomerative approach but with more sharp cluster borders.

\subsection{Variational Autoencoder}
\begin{figure}[!ht]
\centering
\includegraphics[scale=0.29]{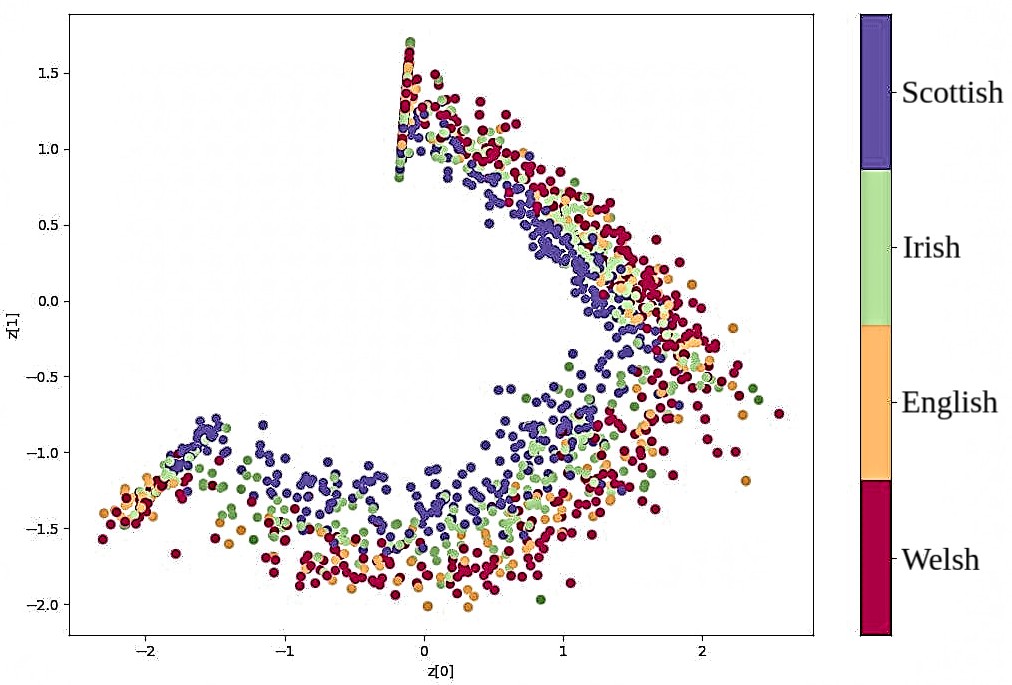}
\caption{VAE latent layer representation}
\label{vae-inner}
\end{figure}

The Variational Autoencoder model was trained on the character encoding of the input data. In contrast to the clustering where we used statistical features, this model is supposed to be able to grasp the sequential nature of the text data.
For feature extraction and analysis, we take the 2-dimensional output of the latent layer of VAE (Figure~\ref{vae-inner}). 

The representation shows an almost symmetrical circular distribution of the features. Although all data points lie close, separation of some language pairs seems possible.
From the plot, we can assume that the least confused languages will be Scottish and Welsh pair. Irish is the most mixed up with the remaining languages and will be hard to differentiate with VAE features. 
VAE feature representation seems to be even more challenging for the classifiers as Irish is mixed with all the remaining languages instead of only Scottish as was the case with the clustering models.

\subsection{Latent Dirichlet Analysis}
\begin{figure}[!ht]
\centering
\includegraphics[scale=0.38]{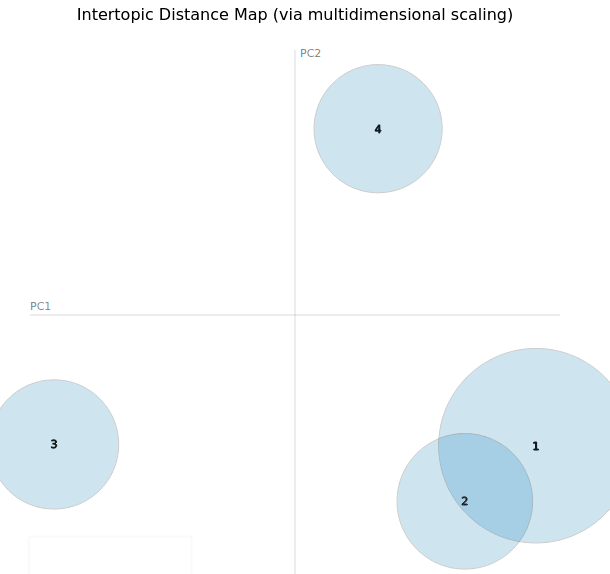}
\caption{LDA topics map}
\label{lda-map}
\end{figure}

The LDA model we created operated on the word n-grams (uni-grams and bi-grams) unlike character n-grams in clustering and character encodings in VAE. Potentially, the topic modelling approach can identify the minimal set of words describing each language.
We explored the relationships between the identified LDA topics. Figure~\ref{lda-map} presents a map of the found word n-gram clusters. As LDA is an unsupervised model, the labels had to be identified separately. We based our language mapping on the each topic's most relevant terms. Table~\ref{lda-top5} presents the top-5 most relevant terms for each topic. However, please note that the topic assignment was performed by a manual analysis of longer lists of the most characteristic terms. The most relevant terms that were found are short words. It is as expected, as the stop words were not removed during the preprocessing, so the LDA model it likely to base its prediction on them. Due to the task's nature, the stop words turned out to be helpful.

\begin{table}
\centering
\caption{LDA top-5 terms per topic}
\begin{tabular}{ p{2cm}p{2cm}p{2cm}p{2cm} } 
  \hline
  Topic 1 (Irish) & Topic 2 (common Celtic words) & Topic 3 (English) & Topic 4 (Welsh and English) \\
  \hline
  agus & an & and & na \\
  an & is & the & yn \\
  na & air & of & ar \\
  do & ach & to & fy \\
  air & ar & for & la \\
  \hline
\end{tabular}
\label{lda-top5}
\end{table}

We discovered that for topics 1 and 3, most of the terms belong to Irish and English, respectively. So these topics represent these languages the most precisely. Such conclusion is helpful when it comes to the separation of English. The topics map shows that English is distant from all other language groupings in the LDA space (topics 3 and 4 on Figure~\ref{lda-map}). 
However, it is not that straightforward with the remaining topics and languages. The topic 2 contains words and bi-grams that simultaneously appear in all considered Celtic languages. 
Such composition of the topic explains its overlap with the topic 1 representing Irish. However, the data records belonging to the topic 2 will be hard to classify to one language.

Finally, the topic 4 includes a combination of words belonging mostly to Welsh and English. 
As English is present in two distinct topics (topics 3 and 4), LDA labels are likely to be a valuable feature for identifying this language. 
The fact that Irish has about a quarter of terms common with other languages explains why it is hard to separate in all of the feature space representation discussed above. Nevertheless, it is likely to be identified comparatively well with LDA.

\section{Results}
\label{sec:results}
\subsection{Supervised Methods}
This section presents results from all our conducted experiments.
We used the standard evaluation metrics such as accuracy and F1. But due to the characteristics of our dataset we used Matthews Correlation Coefficient (MCC) as well. MCC is a more accurate metric for an unbalanced dataset. So it is valuable for our evaluation.
We started with the experiments on the full labelled dataset.

We tested character encodings as the input vectors for the neural models: dense network and CNN. Table~\ref{res-chars} shows that CNN is a definite leader. Therefore, language identification is best performed on small text fragments (CNN filter sizes were set to 2, 3, 4 characters) as opposed to individual character features. This finding supports the choice of the character n-grams as an important feature for our experiments.

\begin{table}
\centering
\caption{Neural models' results on characters encodings}
\begin{tabular}{ lccc } 
  \hline
  Model & Accuracy & MCC & Mean F1 \\
  \hline
  NN & 92\% & 89\% & 92\% \\
  CNN & \textbf{94\%} & \textbf{92\%} & \textbf{94\%} \\
  \hline
\end{tabular}
\label{res-chars}
\end{table}

In the subsequent experiments, we used the character n-grams that are the most popular feature extraction approach for LI (as we showed in the literature review). Two text statistics measures extended them: the average word length and the average number of consonants per word for each sample.
Text statistics had a deteriorating effect on the classification performed by SVM and NN (Tables~\ref{res-svm-full} and~\ref{res-ff-full}, respectively). The NN model using only n-gram features looks especially promising. It has higher results on MCC and F1 for the Irish class, which was identified before as the biggest potential problem. 
The text statistics may not bring the performance improvement in our case as they are indirectly included in the n-grams, i.e., the proportion of consonants can be estimated from the n-grams.
At this stage, we rejected the use of text statistics as a feature extraction method for Celtic LI.

\begin{table}[ht]
\centering
\caption{SVM results summary on test dataset (full train set)}
\begin{tabular}{ lccccc } 
  \hline
  Feature & Accuracy & MCC & Mean F1 & F1 (Irish) & F1 (Scottish) \\
  \hline
  n-gram & \textbf{97\%} & \textbf{96\%} & \textbf{97\%} & \textbf{96\%} & 96\% \\
  n-gram+text stat & 96\% & 95\% & \textbf{97\%} & 95\% & 96\% \\
  \hline
  clusters & 75\% & 68\% & 77\% & 64\% & 51\% \\
  VAE & 32\% & 5\% & 20\% & 0\% & 33\% \\
  LDA & 64\% & 52\% & 65\% & 56\% & 51\% \\
  \hline
  clusters+n-gram & \textbf{97\%} & \textbf{96\%} & \textbf{97\%} & \textbf{96\%} & \textbf{97\%} \\
  VAE+n-gram & \textbf{97\%} & \textbf{96\%} & \textbf{97\%} & 95\% & 96\% \\
  LDA+n-gram & \textbf{97\%} & \textbf{96\%} & \textbf{97\%} & 95\% & 96\% \\
  \hline
\end{tabular}
\label{res-svm-full}
\end{table}  

\begin{table}[ht]
\centering
\caption{NN results summary on test dataset (full train set)}
\begin{tabular}{ lccccc } 
  \hline
  Feature & Accuracy & MCC & Mean F1 & F1 (Irish) & F1 (Scottish) \\
  \hline  
  n-gram & \textbf{98\%} & \textbf{98\%} & \textbf{99\%} & \textbf{98\%} & \textbf{98\%} \\
  n-gram+text stat & \textbf{98\%} & 97\% & 98\% & 97\% & \textbf{98\%} \\
  \hline
  clusters & 76\% & 68\% & 77\% & 64\% & 51\% \\
  VAE & 46\% & 25\% & 36\% & 30\% & 59\% \\
  LDA & 64\% & 52\% & 65\% & 56\% & 51\% \\
  \hline
  clusters+n-gram & 97\% & 96\% & 97\% & 95\% & 96\% \\
  VAE+n-gram & \textbf{98\%} & 97\% & 98\% & 97\% & \textbf{98\%} \\
  LDA+n-gram & \textbf{98\%} & 97\% & 98\% & 97\% & \textbf{98\%} \\
  \hline
\end{tabular}
\label{res-ff-full}
\end{table}

\subsection{Semi-Supervised Methods}
The next step was to evaluate the impact of the unsupervised feature extraction approach. Three types of unsupervised algorithms were used, including clustering, autoencoder, and topic modeling. 
Firstly, the classification models were trained on the whole training set to compare with the performance of the supervised models. Then, only 30\% of the training set was used to assess the possible gains from the unsupervised features when only a small labelled set is available (see Section~\ref{sec:reduced}).

Initial tests were conducted on the whole training set. 
Classification models using clustering output achieved the best performance among the models based on the unsupervised features (Tables~\ref{res-svm-full}, \ref{res-ff-full}).
VAE features led to the weakest results. However, such performance is aligned with the VAE data representation discussed previously. Given such close distribution of all data points, neither SVM nor NN could separate English and Irish data points located in the middle of the data representation (Figure~\ref{vae-inner}).
The models trained on LDA features are generally weaker than those using clustering but more stable than VAE. 

The performance of these models based only on the unsupervised features is not comparable to the supervised models (the best result is 68\% MCC compared to 98\% MCC for the n-gram NN model). Therefore, we decided to test the application of unsupervised features as an extension to the n-gram approach.

We had discovered that the unsupervised features alone are insufficient for high-quality classification. The subsequent experiments involved both unsupervised and n-gram features and were again conducted on the whole dataset. 
The SVM models showed similar results level as the best fully-supervised model, suggesting that n-gram vectors dominated the classification process (Table~\ref{res-svm-full}). However, clustering features were beneficial for the separation of the Irish-Scottish pair outperforming the supervised models on the class-level F1. Moreover, the model with the LDA features has caught up with clustering and improved the classification of Welsh. 

The NN models showed different performance levels. The combination of VAE and LDA features with n-grams outperformed the clustering features. However, it was not clear whether the classifier uses the unsupervised features or its predictions are based solely on the n-grams, given the similarity of the performance results to the supervised models.

\subsection{Experiment with the Reduced Labelled Set}
\label{sec:reduced}

Finally, we evaluated the possibility of the application of the unsupervised features trained on a large amount of unlabelled data to create a classifier trained on a smaller labelled dataset. The previous experiments showed that acceptable performance level is achieved when the unsupervised features are combined with the n-grams. So these models will be compared to the best supervised models that were using only the n-grams on the reduced labelled dataset. In this experiment, 30\% of the training set was used for classifiers, while unsupervised features were prepared on the whole training set without labels; the testing set remained unchanged.

\begin{table}
\centering
\caption{SVM results trained on 30\% of labelled training set, unsupervised features were prepared on the full training dataset}
\begin{tabular}{ lccccc } 
  \hline
  Feature & Accuracy & MCC & Mean F1 & F1 (Irish) & F1 (Scottish) \\
  \hline
  clusters+n-gram & 96\% & 95\% & 96\% & 94\% & 95\% \\
  VAE+n-gram & 96\% & 95\% & \textbf{97\%} & \textbf{95\%} & 95\% \\
  LDA+n-gram & 96\% & 95\% & \textbf{97\%} & 94\% & 95\% \\
  \hline
  only n-gram & 96\% & 95\% & \textbf{97\%} & 94\% & 95\% \\
  \hline
\end{tabular}
\label{svm-30train}
\end{table}

\begin{table}
\centering
\caption{NN results trained on 30\% of labelled training set, unsupervised features were prepared on the full training dataset}
\begin{tabular}{ lccccc } 
  \hline
  Feature & Accuracy & MCC & Mean F1 & F1 (Irish) & F1 (Scottish) \\
  \hline
  clusters+n-gram & 95\% & 94\% & 96\% & 94\% & 95\% \\
  VAE+n-gram & \textbf{98\%} & \textbf{97\%} & \textbf{98\%} & \textbf{97\%} & \textbf{98\%} \\
  LDA+n-gram & 96\% & 95\% & 96\% & 95\% & 97\% \\
  \hline
  only n-gram & \textbf{98\%} & \textbf{97\%} & \textbf{98\%} & \textbf{97\%} & \textbf{98\%} \\
  \hline
\end{tabular}
\label{ff-30train}
\end{table}

When training the classifiers only on the 30\% of the labelled training set, the drop in MCC for the best supervised models was minor. This makes it harder for the models with unsupervised features to show the advantage in performance.

The SVM models showed a similar performance regardless of the features used (Table~\ref{svm-30train}). Only the clustering features appear to be less robust to the dataset reduction. VAE maintains the same performance as the rest of the features, slightly outperforming them on the Irish class F1. 

For the neural network, VAE again achieved the best results among the unsupervised features (Table~\ref{ff-30train}). However, this time, it could not beat the only n-gram supervised model reaching the same results level. Overall, for both classification models, VAE turned out to be the best extension to n-grams on the reduced labelled dataset.

The validity of the addition of the unsupervised features to the n-grams when using the NN as a classifier was questioned above. The confusion matrices in the Table~\ref{ff-30train-cm} will clarify whether the unsupervised features have a positive impact. While the classification of Welsh and English is unchanged with the addition of the features based on the VAE representation, a slight improvement in the separation of the Irish-Scottish pair is observable on the confusion matrices for the models trained on the 30\% of the labelled set. It means that the VAE features are taken into account and have a beneficial impact on the language classes that were the hardest to separate in all our experiments. 
Compared to the n-gram model on the entire dataset, the result is comparable, with only a slight increase in the misclassification of Irish as English (alongside significant labelled set reduction). Therefore, unsupervised features are a reasonable solution to improve a neural network classifier's performance when labelled data is scarce.

\begin{table}
\centering
\caption{Confusion matrices of the best NN models on the 30\% of the labelled training set and the NN n-gram model on the full training set (W -- Welsh, E -- English, I -- Irish, S -- Scottish)}
\begin{tabular}{ clcccc|cccc|cccc } 
  \hline
  &&\multicolumn{12}{c}{Predicted labels} \\
  \hline
  &&\multicolumn{4}{c}{VAE+n-gram (30\%)}&\multicolumn{4}{c}{n-gram (30\%)}&\multicolumn{4}{c}{n-gram (full dataset)} \\
  \hline
  && W & E & I & S & W & E & I & S & W & E & I & S \\
  \hline
  \multirow{4}{*}{\rotatebox[origin=c]{90}{True labels}} & Welsh & 611 & 2 & 0 & 0 & 611 & 2 & 0 & 0 & 611 & 2 & 0 & 0 \\
  & English & 0 & 300 & 0 & 1 & 0 & 300 & 0 & 1 & 1 & 299 & 1 & 0 \\
  & Irish & 2 & 9 & 512 & 11 & 2 & 9 & 510 & 13 & 0 & 3 & 519 & 12 \\
  & Scottish & 0 & 1 & 9 & 528 & 0 & 1 & 11 & 526 & 0 & 1 & 8 & 529 \\
  \hline
\end{tabular}
\label{ff-30train-cm}
\end{table}

\section{Conclusions}
\label{sec:conclusion}

This paper proposed a method to identify the Celtic languages and to differentiate them from English. The prepared SVM and neural classifiers achieved a good performance level that was verified on our collected dataset. We also addressed the issue of the scarce labelled data for the low-resource languages by comparing the results on the reduced and the full labelled training sets.

The dataset collected for this work contained three Celtic languages: Irish, Scottish, Welsh and a smaller English class. The class sizes in the dataset were chosen such as to be sufficient for the task based on the similar previous research. The dataset was collected from the different types of sources as the low-resource languages have limited text availability. Thus, the dataset may be further improved, but it is sufficient to perform the language identification experiments.

Among the neural models, CNN achieved the highest result, albeit showing worse performance than models based on the other feature extraction approaches.
The dense neural model showed a better performance than SVM in all of the experiments.
However, for all classifiers, some language classes posed a continuous challenge. The predictions for the Irish and Scottish classes had the lowest performance. This fact is aligned with the data analysis performed on the output of the unsupervised models.

The proposed approach to use the unsupervised models' output as the feature vectors for classification proved to be effective. And so can be beneficial when the labelled data is scarce. The tests concluded that the unsupervised features are the most successful when used in the combination with the n-gram vectors. 

The experiments on the reduced labelled set proved that the unsupervised features are helpful to minimize the decline in the model's performance with the reduction of the annotated data. The NN model with the autoencoder features outperformed the model using only the n-gram features (Table~\ref{ff-30train-cm}) and achieved MCC close to the result of the model trained on the full dataset; only slight deterioration in the identification of the Irish language was observed. The best model achieved 98\% accuracy and 97\% MCC.

The low-resource languages are challenging due to the scarcity of the available annotated training data. Our results uncovered that the unsupervised feature vectors are more robust to the labelled set reduction. Therefore, they can help to achieve comparable classification performance with much less labelled data. We showed that a neural network operating on the unsupervised and n-gram features could separate even closely-related languages, as has been proved for the Celtic language family.
These results open new opportunities to improve performance and make the corpus annotation process less labor-intensive and expensive. 

\section{Future Work}
\label{sec:future-work}

In our experiments, the pair of languages confused the most was Irish and Scottish. One direction of the future research can aim to solve this issue. Linguistic analysis can establish the discriminating features between these two languages, helping to build a better classifier. This approach is likely to be especially useful when combined with the hierarchical classification. For example, the first level will identify Irish and Scottish as a group, and then these samples will be passed on to a classifier specialized for these languages.

The second direction for the future research revolves around a further investigation of the ways to apply the output of the unsupervised learning model as features for the classifiers. 
Creating a manually annotated high-quality training set may provide a more significant proof of the superiority of the semi-supervised feature learning when the labelled data is scarce.
It would be valuable to extend our experiments with different architectures and parameters of the unsupervised methods.

An interesting research direction is determining how much of training data is needed to solve the task. In our experiments the classifiers performed well even on a relatively small dataset. Knowing how much data is necessary to perform the language identification can be crucial for the low-resource languages.

Finally, another promising research area are the experiments aiming to combine the output of the different unsupervised models into one feature vector. 
Each of the models has shown its strengths and weaknesses. 
If a vector of the integrated features will synthesize each method's strengths, the classification performance will be improved.

\bibliographystyle{unsrtnat}
\bibliography{references}  






\end{document}